# Multi-feature driven active contour segmentation model for infrared image with intensity inhomogeneity

Qinyan Huang[1], Weiwen Zhou[2], Minjie Wan[1,*], Xin Chen[1], Qian Chen[1], Guohua Gu[1,*]

[1]School of Electronic and Optical Engineering, Nanjing University of Science and Technology, Nanjing 210094, China.
[2] Shanghai Institute of Spaceflight Control Technology, Shanghai 201109, China
**Corresponding Authors**
Minjie Wan: minjiewan1992@njust.edu.cn
Guohua Gu: gghnjust@mail.njust.edu.cn

**Abstract:** Infrared (IR) image segmentation is essential in many urban defence applications, such as pedestrian surveillance, vehicle counting, security monitoring, etc. Active contour model (ACM) is one of the most widely used image segmentation tools at present, but the existing methods only utilize the local or global single feature information of image to minimize the energy function, which is easy to cause false segmentations in IR images. In this paper, we propose a multi-feature driven active contour segmentation model to handle IR images with intensity inhomogeneity. Firstly, an especially-designed signed pressure force (SPF) function is constructed by combining the global information calculated by global average gray information and the local multi-feature information calculated by local entropy, local standard deviation and gradient information. Then, we draw upon adaptive weight coefficient calculated by local range to adjust the afore-mentioned global term and local term. Next, the SPF function is substituted into the level set formulation (LSF) for further evolution. Finally, the LSF converges after a finite number of iterations, and the IR image segmentation result is obtained from the corresponding convergence result. Experimental results demonstrate that the presented method outperforms the state-of-the-art models in terms of precision rate and overlapping rate in IR test images.

## 1. Introduction

With the development of smart city technology [1], automatic object detection based on image processing method has achieved great progress in many areas of urban defence, such as pedestrian surveillance [2], vehicle counting [3], public security [4] and so on. Nowadays, IR image segmentation which extracts the object of interest from image plays a fundamental role for all-day object detection and tracking. Among various image segmentation methods [5,6,7,8], ACM has gained popularity because of its excellent ability to obtain closed contours with sub-pixel accuracy [9]. Although, a number of ACMs [10,11,12,13] have achieved satisfactory performances in clear visible images, they only use single image feature information to construct the energy function. Therefore, the results may suffer from false segmentations when the traditional ACMs are directly applied to handle IR images with intensity inhomogeneity, blurred boundary, and low contrast. From the above, it is of great necessity to especially design an effective ACM for IR target segmentation.

Conventional ACMs usually utilize edge information or intensity statistical information to design the energy functional [14]. The edged-based models stop the contour evolution by an edge stopping function (ESF) which is calculated by the gradient information of images. Usually, an ESF has property that the larger gradient is, the smaller value of ESF is. Geodesic active contour (GAC) model [15] is a representative edge-based model, but this model can achieve desirable segmentation results only when the initial contour is set on specific locations, such as surrounding the real boundary. In addition, it is quite sensitive to the interference of noise because the construction of ESF is completely based on gradient information. Jiang et al. [16] improved GAC model by using the gradient vector field and a balloon force so that it can converge accurately near the target boundary. Meanwhile, additive operator splitting (AOS) is introduced to speed up the computing of the model. Wang et al. [17] redefined energy functional by imposing regional information to improve the boundary leaking problem of original GAC model. By using the global average intensity information inside and outside the evolving curve [18], a lot of region-based ACMs have been proposed. Chan-Vese (CV) model [19] represents intensity of object and background by average intensity constants inside and outside contour. In this way, it can segment intensity homogeneity images effectively, but when the objects contain intensity inhomogeneity, false segmentations will occur. To make up for the deficiencies of the CV model, local intensity information is imposed to construct the LSF. The local binary fitting (LBF) model, proposed by Li et al. [20], uses Gaussian kernel function to construct local energy function. Related experiments show that LBF model outperforms CV model in handling intensity inhomogeneous IR images, however it is prone to local minimum and is sensitive to initialization. For faster and more accurate image segmentation, Ding et al. [21] proposed local pre-fitting (LPF) model in which local average image intensities are used to form two pre-fitting functions. Because pre-fitting functions are calculated before the curve evolution, they are invariant during the iterations. By this means, the computational time of LPF model decreases remarkably. Local region-based Chan–Vese (LRCV) model developed [22] constructs the energy functional using local average intensity information in order to extract objects with intensity inhomogeneity correctly. Wang et al. [23] designed entropy weighted fitting (EWF) model, in which Kullback Leibler divergence is introduced as a measuring metric for comparing the input image and three local fitting images and inhomogeneity entropy descriptor is regard as weight to adjust afore-discussed images. Laplacian of Gaussian (LoG) which is essentially a second-order differential operator can detect the edge of image effectively while region-scalable fitting (RSF) which imposes kernel function to estimate local intensity is a local region-based model. By combining the advantage of these two terms, Ding et al. [24] put forward LoG & RSF model to remedy the shortcoming of sensitivity to contour initialization of RSF model and achieves more accurate segmentation results in related experiments. However, the LoG operator is sensitive to the change of intensity, when the boundaries of images are blurred and background regions are complex, some false edges will still occur. Another strategy to cope with inhomogeneity is to combine ACMs with bias field estimation. Li et al. [25] proposed local intensity clustering (LIC) model via replacing the average intensity constant with a local cluster function. The intensity inhomogeneous regions can be corrected by the estimated bias field. But the clustering variance is not taken into consideration in this model and it is time-consuming. Locally statistical active contour model (LSACM) [26] can be categorized into K-means clustering method. It defines a clustering criterion to combine pixels of the same class effectively and is able to obtain a complete segmentation contour. Huang et al. [27] designed a region-based SPF function in order to handle inhomogeneous images correctly. Furthermore, the kernel function used for regularizing LSF is constrained by an adaptive scale parameter and the bias field is initialed in a new way, so this model can estimate the bias field more accurately and the contour can be initialized in arbitrary locations. Hai et al. [28] put forward a region-bias fitting (RBF) model to deal with inhomogeneous images by imposing a constraint term using region bias. By using local difference between the input image and the bias estimated image, Shan et al. [29]. developed local region-based fitting (LRBF) model in which neighborhood of every pixel constructs a local energy function and these functions are then integrated to form total energy functional. Furthermore, there are some other ACMs. Zhang et al. [30] designed a novel model named the selective local or global segmentation (SLGS) model, which utilizes statistical information to form a SPF function and then substitutes the ESF of GAV model by the new SPF function. Mukherjee et al. [31] put forward an edge independent segmentation model which estimates the object and background intensity by a series of Legendre polynomial functions and it is robust to the change of grayness. However, IR images have more complex properties e.g. intensity inhomogeneity, blurred boundary, and low contrast and so on [32]. Above-mentioned ACMs still cannot deal with IR target perfectly, so it is essential to design a special ACM for IR image segmentation.

In this paper, we propose a new ACM model by constructing SPF functions with both global and local information. The global term is to control the evolving curve to pass through the flat background region quickly and to prevent the evolving curve falling into local minimum, so we choose global average intensity information to form the global term. The local term is made up of the information of entropy, standard deviation and gradient, which accurately guides the convergence of the evolving curve when it is close to the real edge. At the same time, we calculate the adaptive weight coefficient based on local range information in order to incorporate the global term and local term. Then, the afore-mentioned SPF function is used to construct the LSF, and Gaussian filter is also utilized to regularize the LSF in order to remove the re-initialization process. After a finite number of iterations, LSF tends to converge, and its zero LSF corresponds to the contour boundary of object, i.e., the IR target can be thus completely segmented.

The main contributions of this paper can be summarized as follows:
(1) Average intensity information inside and outside the contour is combined through Heaviside function to form the global term, which prevents the curve evolution falling into local minimum.

(2) Local multi-feature information, similar to entropy, standard deviation and gradient, is introduced to construct a new SPF function for overcoming the interference of intensity inhomogeneity;

(3) An adaptive weight coefficient calculated by local range information is proposed to guide the evolving curve more effectively.

The rest of this work is organized as below. The conventional SLGS model is introduced in Section 2 and we will discuss the reason why this model is unable to segment the infrared images correctly. Section 3 introduces the theory of our proposed model in detail, especially the construction method of the new SPF function. Related comparative experimental are shown in Section 4 and a brief summary is drawn in Section 5.

## 2. Motivation

Zhang et al. [30] assumed that image intensities are statistically homogeneous and they constructed an SLGS model. As shown in Eq. (1), the LSF $\phi$ is initialized to a constant and the sign of the constant is opposite inside and outside the contour.

$$\phi((u,v), t=0) = \begin{cases} -c_0, (u,v) \in \Omega_0 - \partial\Omega_0 \\ 0, (u,v) \in \partial\Omega_0 \\ c_0, (u,v) \in \Omega - \Omega_0 \end{cases}, \tag{1}$$

where constant $c_0 > 0$, $\Omega$ denotes the input image and $\Omega_0$ represents a subset of the $\Omega$; The zero level of $\phi$ corresponds to the boundary of $\Omega_0$ and it is described as $\partial\Omega_0$. Then, the model **uses** the average intensity information to form a SPF function and its values are normalized in the range [-1, 1]. The role of the SPF function is to determine the evolution direction. More specifically, it shrinks when the curve is outside of the object and expands when the curve is inside of the object. The SPF function is defined as:

$$spf(I(u,v)) = \frac{\text{Img} - \frac{c_1 + c_2}{2}}{\max \left| \text{Img} - \frac{c_1 + c_2}{2} \right|}, \tag{2}$$

where $c_1$ and $c_2$ serve as the average intensities inside and outside the contour, respectively. Their mathematical expression as follows:

$$c_1(\phi) = \frac{\int_\Omega I(u,v) \cdot H(\phi) dx}{\int_\Omega H(\phi) dx}, \tag{3}$$

$$c_2(\phi) = \frac{\int_\Omega I(u,v) \cdot (1 - H(\phi)) dx}{\int_\Omega (1 - H(\phi)) dx}, \tag{4}$$

where $H(\phi)$ means the regularized Heaviside function and it is defined as follows:

$$H_\varepsilon(\phi) = \frac{1}{2}\left(1 + \frac{2}{\pi}\arctan\left(\frac{\phi}{\varepsilon}\right)\right), \tag{5}$$

where $\varepsilon$ is control factor; $H(\phi) = 1$ when $\phi > 0$, otherwise $H(\phi) = 0$. The LSF of

SLGS model can be written as:

$$\frac{\partial \phi}{\partial t} = spf\left(I(u,v)\right) \cdot \alpha \cdot |\nabla \phi|. \tag{6}$$

As shown in the first row of Fig1, SLGS model can obtain desirable segmentation result when the image intensities are statistically homogeneous in visible image. However, when SLGS model is applied to IR image with intensity inhomogeneity (see the second row of Fig1), its segmentation precision significantly decreases. This is because SLGS model assumes the object and background are homogeneity and only uses single average intensity constants to present the image intensity. However, the object always contains intensity inhomogeneity, so the object cannot be extracted exactly.

To overcome the afore-discussed deficiency of SLGS model, local feature information is introduced to guide the curve evolution in inhomogeneous regions. A special term which contains both the global intensity information and the local feature information are designed to replace the original SPF function. In this way, the new ACM can extract IR objects with intensity inhomogeneity effectively.

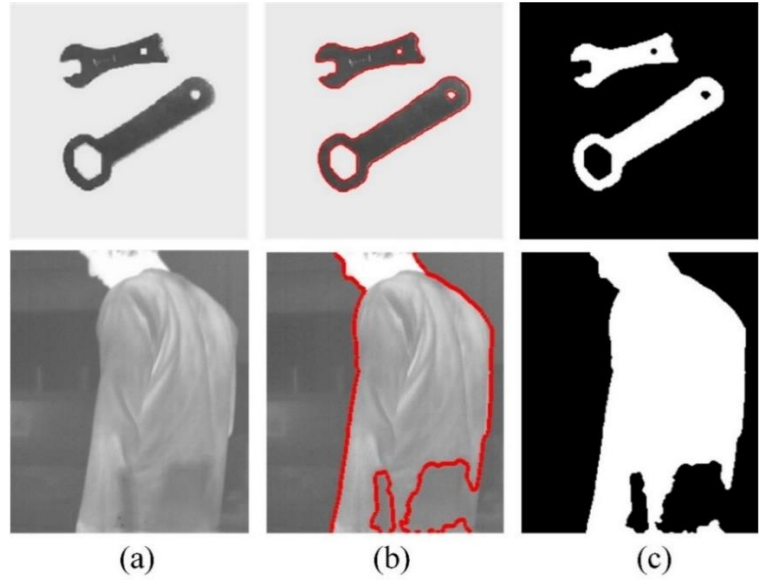

Fig.1 Segmentation results of SLGS model: (a) original image;
(b) segmentation result; (c) binary segmentation result.

## 3. Theory

According to the analysis in Section 2, we know that SPF function which only contains global gray average information cannot segment IR image effectively. The key technique of this paper is to design a new SPF function by combining global gray average information with local multi-feature information. Multi-features information is introduced to highlight the significant difference in those regions with intensity inhomogeneity so that the evolution direction of curve can be determined more accurately. By the guidance of the new SPF, the presented ACM is able to segment intensity homogeneous IR images correctly. The structure diagram is shown in Fig.2.

## 3.1 Design of global term

The global term aims to prevent the curve falling into local minimum and drive the curve to pass through the flat background region quickly are the design principles of the global term. Hence, for a given IR image $I$, we utilize global gray average information to form the global term and its calculation formula is described as:

$$SPF_{global}(u,v) = I(u,v) - (c_1 \cdot H(\phi) + c_2 \cdot (1 - H(\phi))), \qquad (7)$$

where $c_1$ and $c_2$ are calculated by Eq. (3) and (4), respectively. The second term of Eq. (7) represents the fitting image and it can be constructed by combining the average intensities inside with outside the contour through Heaviside function $H(\phi)$. When the difference between the original image and the fitting image is the smallest, the $SPF_{global}$ function tends to 0 and the curve stops evolving. Furthermore, the sign of $SPF_{global}$ function can determine the direction of pressure forces, so that the evolving curve can shrink or expand to extract the object.

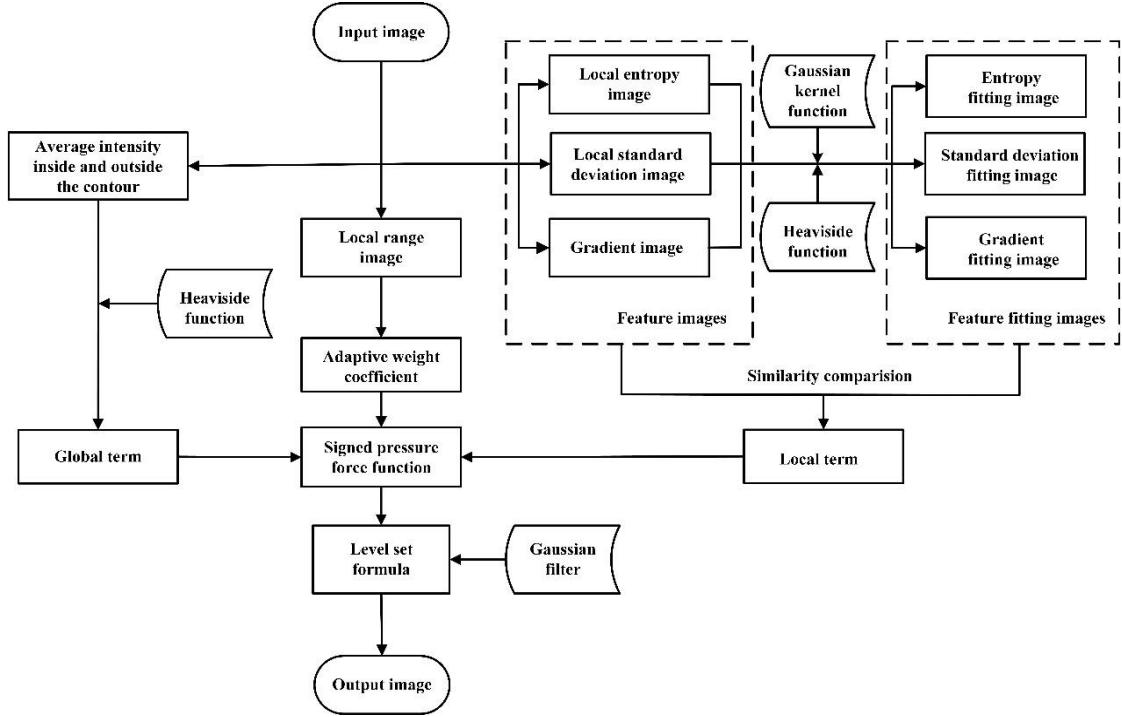

Fig.2 Structure diagram of the proposed method.

## 3.2 Design of local term

We consider that it is inadequate to represent IR image with intensity inhomogeneity only using global gray average information. To determine a more precise evolution direction, local entropy, local standard deviation and gradient information are employed to construct the local term. We set a local window $W_n$ whose size is $n \times n$ surrounding the center pixel $(u,v)$, then the entropy image $I\_en$, standard deviation image $I\_std$ and gradient image $I\_grad$ are computed by the following formulas:

$$I\_en(u,v) = \sum_{i=0}^{L} p_i \log p_i, \left(p_i = \frac{n_i}{n \times n}\right), \tag{8}$$

$$I\_std(u,v) = \sqrt{\frac{1}{n \times n} \sum_{(u,v) \in W_n} (I(u,v) - \mu)^2}, \tag{9}$$

$$I\_grad(u,v) = \sqrt{(I(u+1,v) - I(u,v))^2 + (I(u,v+1) - I(u,v))^2}, \tag{10}$$

where $L$ refers to the total number of gray levels in the local window; $i$ means a certain gray value and $n_i$ is the number of pixels corresponding to $i$; $p_i$ is the probability of pixels with gray value $i$ in the window; $\mu$ means the average gray level of all pixels in $W_n$ and its calculation formula is $\mu = \frac{1}{n \times n} \sum_{(u,v) \in W_n} I(u,v)$. According to previous works, the gray change in the region of IR object is more significant than background region. The gray change may cause a tremendous variation in local entropy. Thus, local entropy is used to represent the change of texture information. Similarly, the standard deviation and gradient are utilized to reflect the change of local intensity. When the values of standard deviation and gradient are large in the local window, it indicates that this region may be the edge of object. That is to say, the selected features describe the local property in a comprehensive way.

According to LBF model [20], a Gaussian kernel function $K_\sigma$ is introduced to embed the local intensity information, and the approximate local intensity calculation formulas inside and outside the contour are shown in Eq. (11) and (12).

$$f_{in}(u,v) = \frac{K_\sigma * [M_1(\phi) I(u,v)]}{K_\sigma * M_1(\phi)}, \tag{11}$$

$$f_{out}(u,v) = \frac{K_\sigma * [M_2(\phi) I(u,v)]}{K_\sigma * M_2(\phi)}, \tag{12}$$

where $\sigma$ is standard deviation and $*$ represents the convolution operator; $M_1(\phi) = H(\phi)$, $M_2(\phi) = 1 - H(\phi)$ and $H(\phi)$ is given by Eq. (5). $f_{in}(u,v)$ and $f_{out}(u,v)$ denote approximate image intensities in the neighborhood of pixel $(u,v)$ inside and outside the evolving curve respectively. On this basis, we also impose a Gaussian kernel function $K_\sigma$ to embed the local entropy, local standard deviation and gradient information. Their mathematical expressions are as follows:

$$\Sigma_{in}(u,v) = \begin{cases} entro_{in}(u,v) = \dfrac{K_\sigma * [M_1(\phi) I\_en(u,v)]}{K_\sigma * M_1(\phi)} \\ std_{in}(u,v) = \dfrac{K_\sigma * [M_1(\phi) I\_std(u,v)]}{K_\sigma * M_1(\phi)} \\ grad_{in}(u,v) = \dfrac{K_\sigma * [M_1(\phi) I\_grad(u,v)]}{K_\sigma * M_1(\phi)} \end{cases}, \tag{13}$$

$$\Sigma_{out}(u,v) = \begin{cases} entro_{out}(u,v) = \dfrac{K_\sigma * [M_2(\phi) I\_en(u,v)]}{K_\sigma * M_2(\phi)} \\ std_{out}(u,v) = \dfrac{K_\sigma * [M_2(\phi) I\_std(u,v)]}{K_\sigma * M_2(\phi)} \\ grad_{out}(u,v) = \dfrac{K_\sigma * [M_2(\phi) I\_grad(u,v)]}{K_\sigma * M_2(\phi)} \end{cases}, \quad (14)$$

where $\Sigma_{in}(u,v)$ contains three local feature information inside the contour and $\Sigma_{out}(u,v)$ contains three local feature information outside the contour. Every feature information from both side of contour can be combined by Heaviside function to construct the single feature fitting image. The similarities are measured by calculating the distance between feature fitting images and original feature images. Thus, we can obtain three SPFs dominated by three local features respectively, which are defined as Eq. (15)-(17):

$$spf\_en(u,v) = I\_en(u,v) - [entro_{in}(u,v) \cdot M_1(\phi) + entro_{out}(u,v) \cdot M_2(\phi)], \quad (15)$$

$$spf\_std(u,v) = I\_std(u,v) - [std_{in}(u,v) \cdot M_1(\phi) + std_{out}(u,v) \cdot M_2(\phi)], \quad (16)$$

$$spf\_grad(u,v) = I\_grad(u,v) - [grad_{in}(u,v) \cdot M_1(\phi) + grad_{out}(u,v) \cdot M_2(\phi)], \quad (17)$$

where $spf\_en$ and $spf\_std$ represent the driving forces provided by entropy information and standard deviation information, and the gradient driving force is determined by $spf\_grad$. Then, the local term is computed as the sum of three functions:

$$SPF_{local}(u,v) = spf\_en(u,v) + spf\_std(u,v) + spf\_grad(u,v). \quad (18)$$

By calculating Eq. (18), the local feature information is employed to provide pressure force and the evolution direction of every pixel on the curve can be thus determined.

### 3.3 Adaptive weight coefficient

In order to combine the afore-mentioned global and local term in our model more reasonably, the adaptive weight coefficient $\omega(u,v)$ is proposed to construct a complete SPF as follows:

$$SPF_{total} = \omega(u,v) \cdot SPF_{global} + (1 - \omega(u,v)) \cdot SPF_{local}, \quad (19)$$

where $SPF_{total}$ should be normalized to the range [-1,1]. As shown in Eq. (19), the value of $\omega(u,v)$ decides whether the SPF function is mainly driven by global term or local term. On the one hand, when the intensity of image region changes remarkably, the global gray average information cannot represent the local intensity change and may result in false segmentation. At this point, local feature information is essential to be taken into consideration, i.e., the weight of local term $SPF_{local}$ should be increased. On the other hand, when the intensity varies smoothly, it indicates that the contour is in the background region, so we should increase the weight of global term $SPF_{global}$ in order to pass through the background region quickly. Hence, we choose local range to quantificationally reflect intensity variation, the mathematical formula of which can be written as:

$$range(u,v) = \max(W_m(u,v)) - \min(W_m(u,v)). \tag{20}$$

The significance of Eq (20) is to compute the difference between maximum and minimum in the $m \times m$ sized local window $W_m$. According to the above-mentioned analysis, the weight coefficient $\omega(u,v)$ is designed as follows:

$$\omega(u,v) = \frac{1}{1 + \lambda \cdot range(u,v)}, \tag{21}$$

where $\omega \in (0,1)$ and $\lambda$ is a constant. If the evolving curve is in flat region without any objects, the value of local range is small, i.e., $\omega$ tends to 1, so the global term dominates the curve evolution. While the evolving curve is near the real boundary, the value of local range becomes large and $\omega$ tends to 0. At this time, curve evolution is mainly controlled by the local term. By this means, the weight of global term and local term can be adaptively determined by the new SPF function, leading to the advantage that our ACM can segment IR image with intensity inhomogeneity more accurately.

### 3.4 Implementation

By exploiting the new SPF function, the LSF can be rewritten as follows:

$$\frac{\partial \phi}{\partial t} = SPF_{total} \cdot \alpha \cdot |\nabla \phi| = \left[w \cdot SPF_{global} + (1-w) \cdot SPF_{local}\right] \cdot \alpha \cdot |\nabla \phi|. \tag{22}$$

In traditional ACMs, Signed Distance Function (SDF) is used for the initialization of LSF and it means the shortest distance from a pixel to the evolving curve. Thus, when the location of evolving curve has changed, the values of SDF also need to be updated, i.e., the process of re-initialization is necessary during the iterations. In order to reduce the computational time of SDF and re-initialization, in our model, the LSF $\phi$ is initialized to constant as Eq. (1) and a selective step is utilized to penalize LSF to be binary. Whether the selective step is performed depends on the desired property. If we only want to extract the object of interest, it is necessary; otherwise, it is unnecessary. Then Gaussian function is utilized to regularize the LSF after each iteration as:

$$\phi_t = \phi_{t-1} * G_{\sigma\_\phi}, \tag{23}$$

where $G_{\sigma\_\phi}$ represents a Gaussian function. $\phi_{t-1}$ means the evolution result of (t-1)-th iteration and $\phi_t$ can be considered as the initial state of t-th iteration. By executing Eq. (23), the procedure of re-initialization can be removed.

In order to stop evolution in time, we set a stopping criterion of curve evolution as follow:

$$|\phi_t - \phi_{t-1}| < \delta, \tag{24}$$

where $\delta$ is convergence threshold and the Eq. (24) means the LSF does not change obviously any further.

To describe the implementation of our model clearly, the main procedures of the proposed algorithm are summarized as follows:

---
**Algorithm 1**
---
**Input:** an IR image $I$
1. initialize the LSF $\phi$ to be a binary function according to Eq. (1).
2. **While** not convergence **do**
3. **For** every pixel **do**
4.     Construct the global term $SPF_{global}$ according to Eq. (7);
5.     Construct the local term $SPF_{local}$ according to Eq. (18);
6.     Compute the adaptive weight coefficient $\omega$ using Eq. (21).
7. **End**
8. Construct complete SPF function using Eq. (19).
9. Update the LSF according to Eq. (22).
10. Regularize the LSF according to Eq. (23).
10. **If** $|\phi_t - \phi_{t-1}| < \delta$
11.     break;
12. **End if**
13. **End**

**Output:** the zero-level set of $\phi$: $\phi((x,y),t) = 0$.

---

## 4. Experimental results

Both qualitative and quantitative experiments are conducted with real IR images to demonstrate the accuracy and efficiency of our method in this section. All the experiments are implemented using Matlab 2016a on a PC with Intel Core i7-8565U, CPU 1.80GHz and RAM 8.0 GB.

### 4.1 Parameter Setting

According to previous works and relative experimental results, all related parameters as well as their meaning and default value in our experiments are listed in Table.1.

Referring to LBF model [20], the standard deviation of Gaussian function $K_\sigma$ utilized for embedding local feature information is $\sigma = 3.0$. The size of local window used for computing local features is fixed to be $9 \times 9$. LSF is initialized to constant $\rho$ and we set $\rho = 1$ as SLGS model. We use Gaussian filter whose standard deviation is $\sigma\_\phi$ to re-initialize LSF. If the $\sigma\_\phi$ is set too small, the interference of noise may influence the segmentation result. While the $\sigma\_\phi$ is too large, it may cause inaccurate boundary detection. $\sigma\_\phi$ ranges from 0.8 to 1.5 by experience and it set to be 1.0 in our experiments. We apply convergence threshold $\delta = 10^{-5}$ as a criterion to stop the curve evolution. The parameters $\lambda$ that determines the weight of global term and local term in Eq. (21) and $\alpha$ that determines the convergence rate of the LSF in Eq. (22) influence the precision rate of segmentation significantly. Further sensitivity analysis about $\lambda$ and $\alpha$ are presented in follow.

Table.1 Initialization of Parameters

| Parameter | Meaning | Setting Value |
|---|---|---|
| $c_0$ | The constant used for initializing the LSF | 1.0 |
| $\sigma$ | The standard deviation of the Gaussian kernel function used for embedding local features | 3.0 |
| n | The width of local window to calculate the feature information | 9 |
| m | The width of local window to calculate the adaptive coefficient | 5 |
| $\lambda$ | Determine the weight of global term and local term | 0.5 |
| $\alpha$ | The balloon force of LSF | 400 |
| $\sigma\_\phi$ | The standard deviation of the Gaussian filter used for regularizing LSF | 1.0 |
| $\delta$ | The convergence threshold | $10^{-5}$ |

To obtain suitable settings of $\alpha$ and $\lambda$, we introduce F value to make sensitivity analysis in a quantitative way and it is defined as follows:

$$F = \frac{2 \cdot \text{Pre} \cdot \text{Rec}}{\text{Pre} + \text{Rec}}, \quad (25)$$

where $F \in [0,1]$; *Pre* represents the precision rate and *Rec* represents recall rate, whose calculational formulas are given as:

$$\text{Pre} = \frac{I_{cor}}{I_{out}}, \quad (26)$$

$$\text{Rec} = \frac{I_{cor}}{I_{truth}}, \quad (27)$$

where $I_{out}$ denotes the output binary image segmented by the ACMs; $I_{truth}$ represents the binary image of the ground truth; $I_{cor}$ can be computed by following formula: $I_{cor} = I_{truth} \cap I_{out}$ and it means the correctly segmented region. Generally speaking, the larger F values is, the more accurate segmentation result is.

We select three types of IR images to make the sensitivity analysis about $\alpha$ and $\lambda$. These images all contain a degree of local intensity inhomogeneity, but they represent images with single object, images with multiple objects and images with low resolution, respectively. While other parameters are fixed, we set $\lambda$ changing from 0 to 3(the step length is set as 0.5) and the F value is calculated for each $\lambda$. The experimental result is drawn in Fig.3(a). These $\lambda$-F value curves indicate that along with the increase of $\lambda$, the F values tend to decrease. Focusing on the average curve (marked in red), we consider that the F values are acceptable when $\lambda$ varies in the range [0,1]. It reaches the maximum when $\lambda = 0.5$, so we set $\lambda = 0.5$ for our experiments. As for $\alpha$, it changes from 100 to 800 and the step length is chosen as 100. The $\alpha$-F value curves drawn according to experimental result are shown in Fig.3(b). We can discover that when $\alpha$ varies from 100 to 400, the F values increase remarkably and when $\alpha > 400$, the $\alpha$-F value curves are almost a straight line with high F values. Considering the segmentation efficiency, we adopt $\alpha = 400$ in our experiments finally. Because of the difference of IR images, the setting values listed in Table.1 are just reference values and they still need to be adjusted for obtaining more accurate segmentation result.

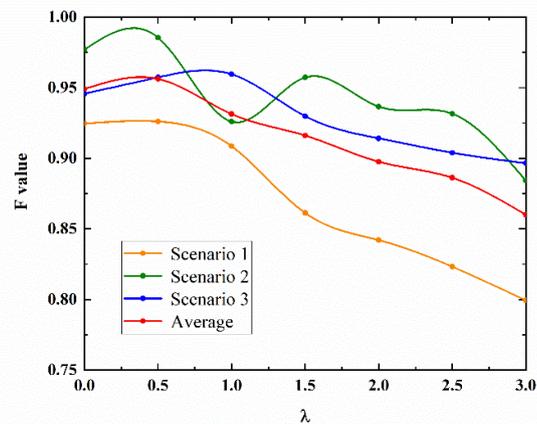

(a) $\lambda$-F value curves.

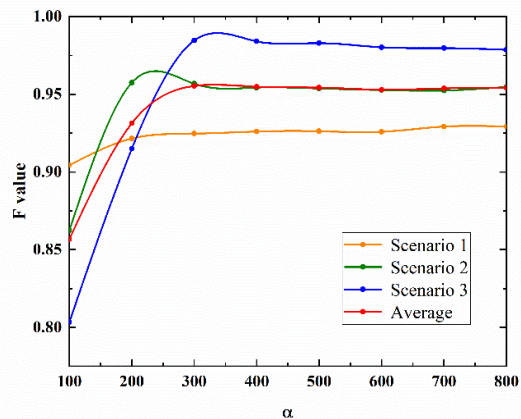

(b) $\alpha$-F value curves.
Fig.3 Sensitivity analysis

### 4.2 Comparative experiment

In order to compare our proposed method with other ACMs in qualitative and quantitative ways, 16 IR images which all contain a degree of inhomogeneity are selected to make comparative experiments. In particular, IR.1-IR.11 only contain single object and others are more than one. We compare the result of our method with 6 well-known ACMs: GAC [15], CV [19], SLGS [30], LPF [21], LoG & RSF [24] and LIC [25] model which all have been discussed in Section2.

### 4.2.1 Comparison of segmentation result

As shown in Fig.4, the original images are presented in the first column and we set uniform initial contour whose size is $40 \times 40$ in the image center for the ACMs (marked in green). Corresponding to columns (2-9), the segmentation results of GAC,

CV, SLGS, LPF, LoG & RSF, LIC, proposed method and the ground truth are shown respectively, in which the finally evolving curves are marked in red.

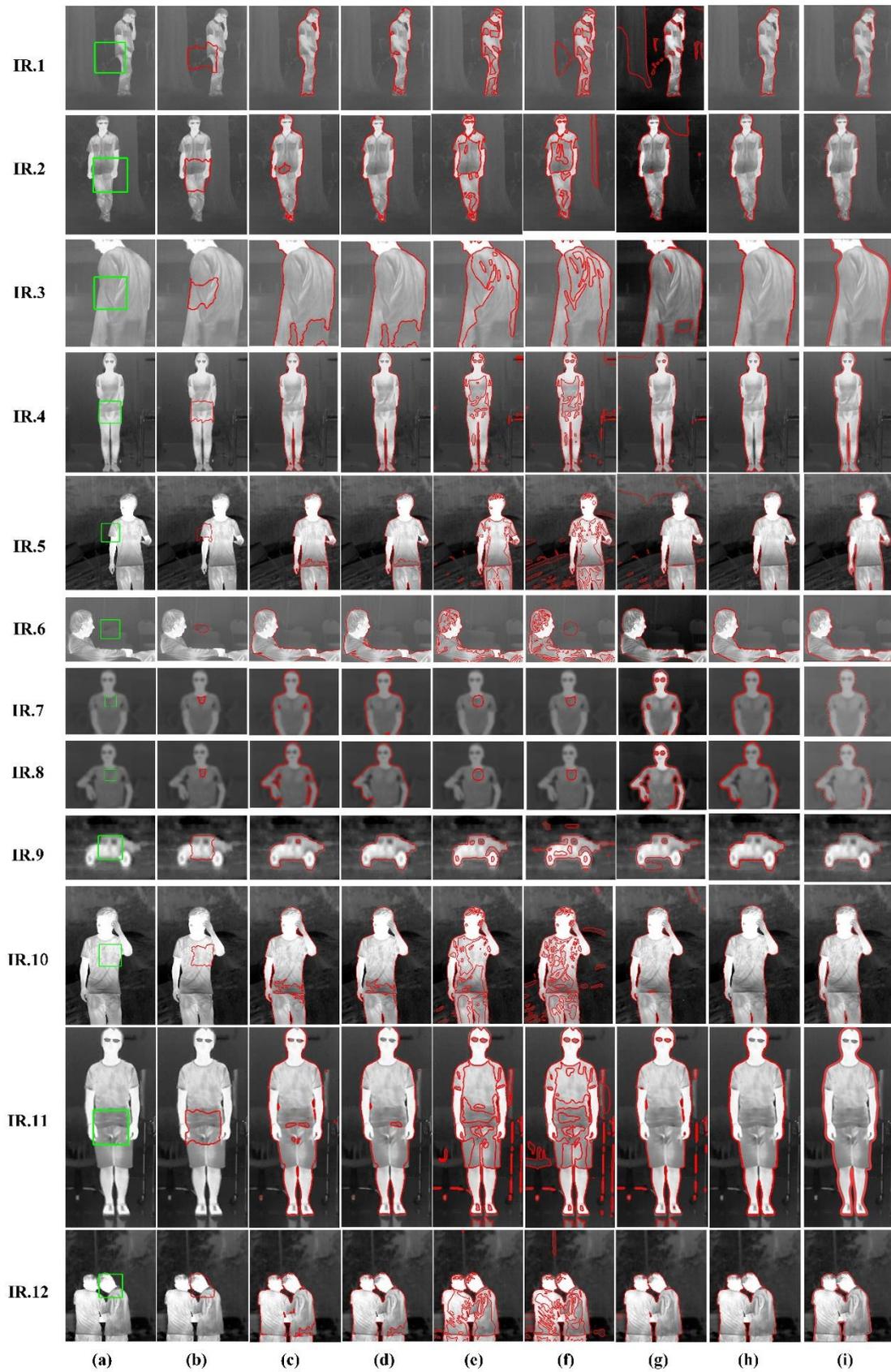

Fig.4 Cont.

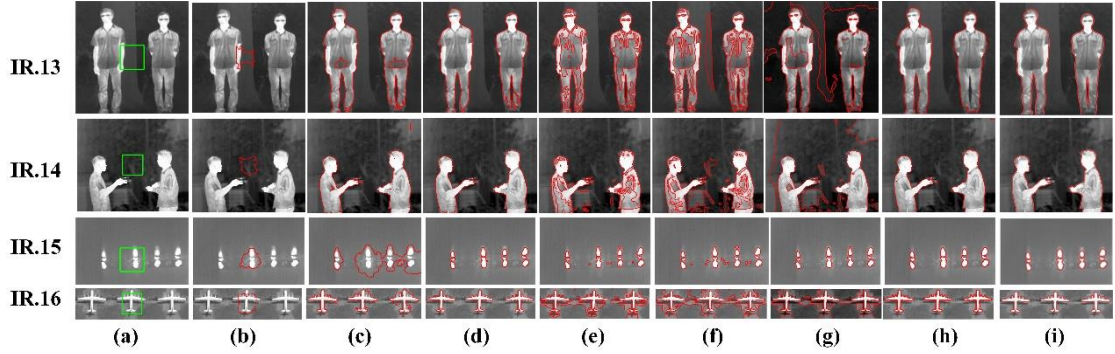

Fig.4 Segmentation results of ACMs: (a) initial contour; (b) GAC;(c) CV (d) SLGS; (e) LPF; (f) LoG & RSF; (g) LIC; (h) the proposed method; (i) ground truth.

We find that GAC model does not converge on all the test images due to its sensitiveness to the initialization of contour. According to Zhang's research [30], the object is able to be segmented correctly, only when the initial contour is set in the appropriate location. But in our experiments, contours are initialized in the images center, which does not satisfy this condition, so the segmentation results of GAC model are unsatisfactory. CV model assumes that intensities of object and background are statistically homogeneous and it only utilizes global average intensity constants to construct the energy functional. However, IR images always involve a degree of inhomogeneity. Hence, CV model always causes false segmentations in clothes region with intensity inhomogeneity. SLGS model combines the advantages of GAC and CV, thus it can handle IR images with blurred boundary and low contrast effectively (see IR.9 and IR16). Nevertheless, when the local intensity inhomogeneity of object is serious, SLGS model has the same drawback as CV model, i.e., false segmentation and boundary leakage will occur in those inhomogeneous areas.

LPF model and LoG & RSF model are representative models of local fitting-based ACMs. Energy functions of local fitting-based ACMs usually involve the local intensity information but ignore the global information, so they are sensitive to the state of initialization and always fall into local minima due to lacking global information. According to the segmentation results of LPF model and LoG & RSF model, we can verify the above-mentioned conclusion that they both fall into local minima, so the curve always stops evolving in the interior of object with a number of false segmentations. In particular, when images contrast is low (such as IR7, IR8 in Fig.4), the active contours of LPF model and LoG & RSF model almost stop evolving during the whole iterative procedure.

LIC model estimates the intensity in a neighborhood of each pixel by designing a local clustering criterion function and then integrates the local function to construct the total energy functional. Meanwhile, the bias field estimation is introduced to intensity inhomogeneity correction. Because the clustering variance is not taken into consideration, as we can see from the segmentation result, this model still cannot handle IR image region with serious intensity inhomogeneity effectively. In addition, this model is sensitive to the change of intensity, so we can find that the contours do not converge in those background regions with intensity change (see IR1, IR2, IR5, IR13,

IR14).

Compared with the conventional ACMs, our method performs more precisely and robustly in segmenting the test images. Note that our method can handle the IR.4 and IR.11 which are contaminated by noise effectively. Comparing with other ACMs, our global term makes curves avoid falling into local minima, and the local term takes local feature information into consideration to provide more accurate evolving direction in the region with intensity inhomogeneity. By this means, we can finally obtain complete and correct contours.

### 4.2.2 Comparison of segmentation accuracy

In this section, F values are calculated to evaluate the segmentation results in a quantitative way. As shown in Fig.5, the binary segmentation results corresponding to above-mentioned 6 ACMs, proposed method and the ground truth are presented.

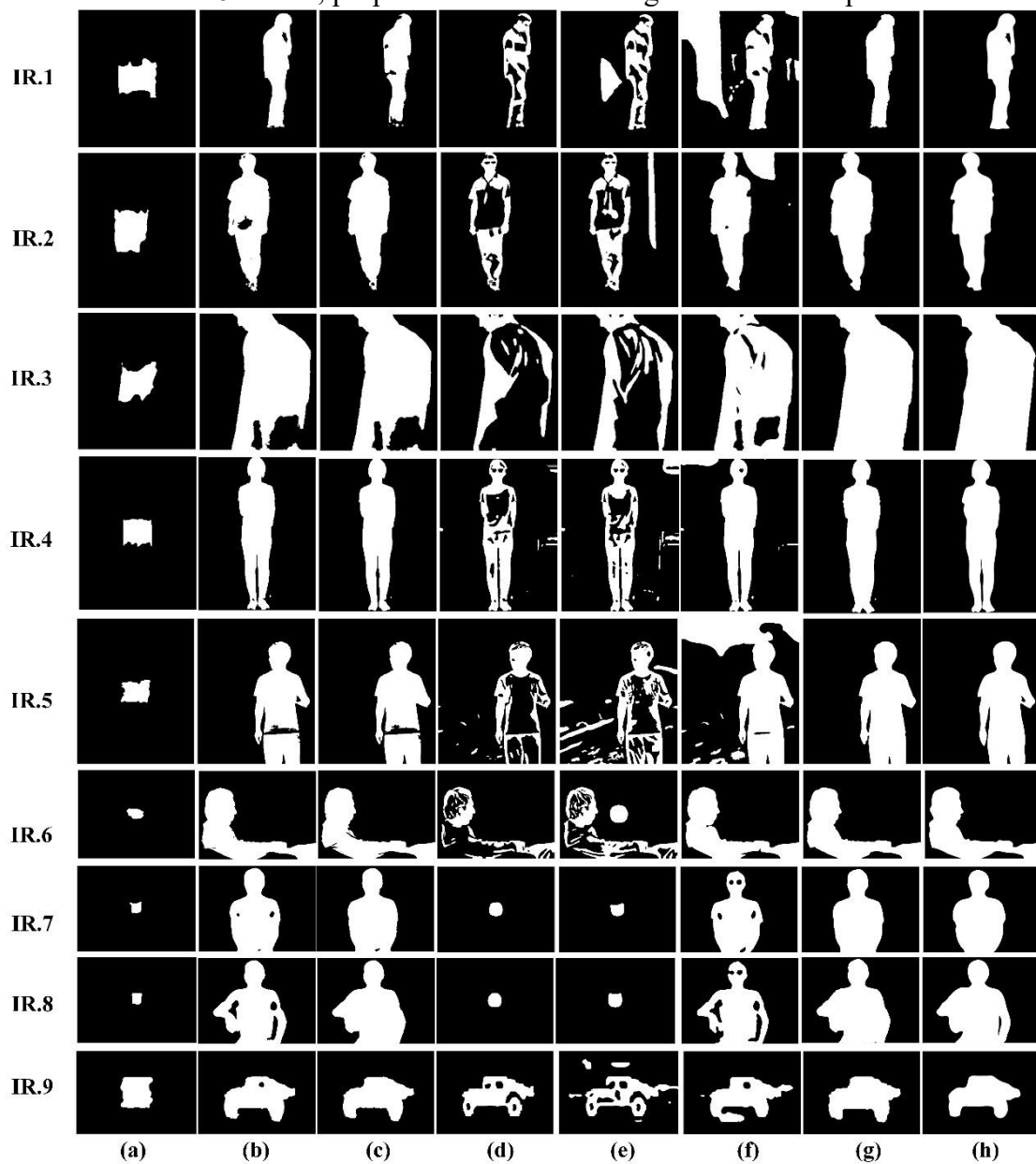

Fig.5 Cont.

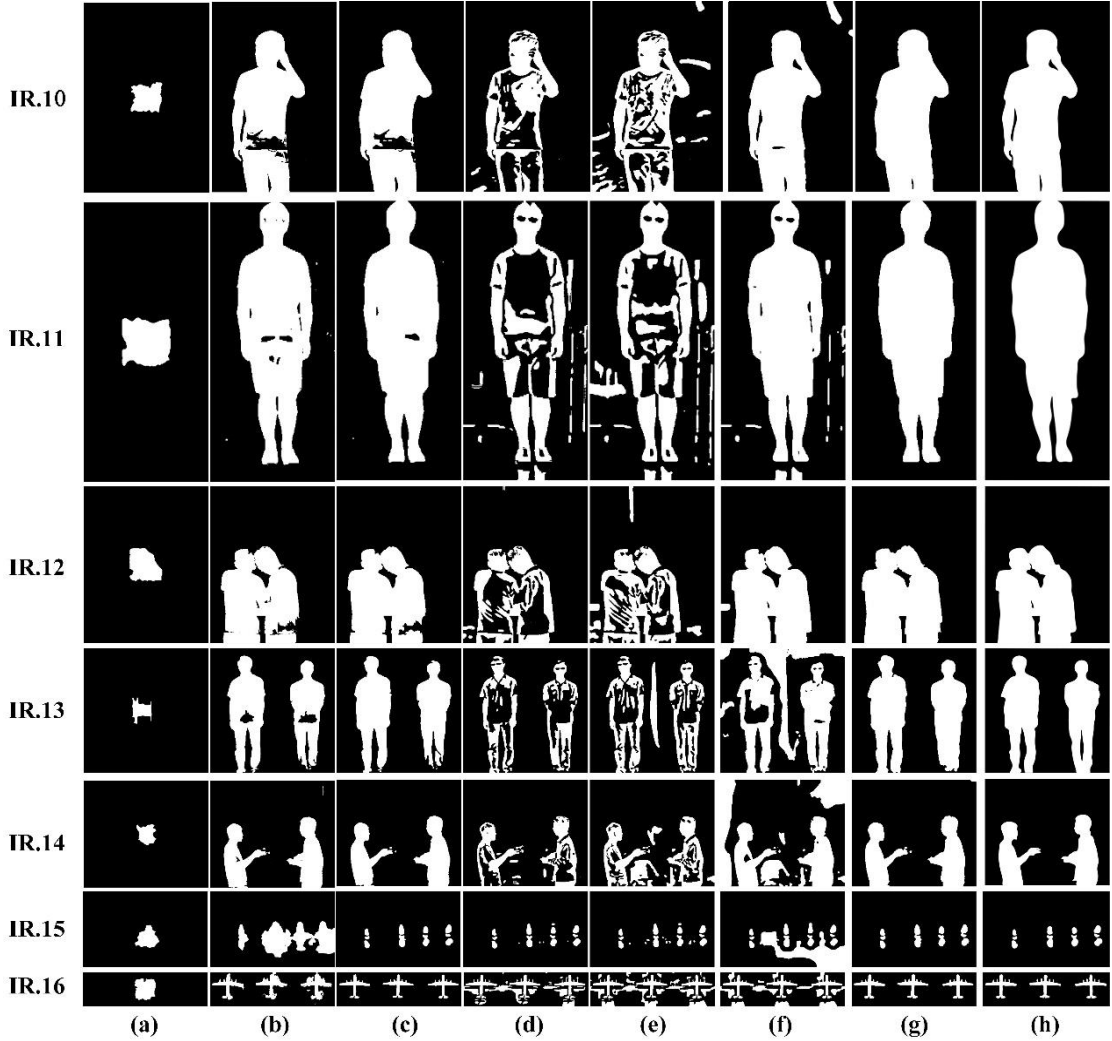

Fig.5 Binary segmentation results of ACMs: (a) GAC; (b) CV; (c) SLGS; (d) LPF;
(e) LoG & RSF; (f) LIC; (g) the proposed method; (h) ground truth.

The F values of 16 IR images segmented by 6 traditional ACMs and our method are listed in Table.2, and the three top largest F values are marked in red, green and blue, respectively.

From Table.2, we find that the F values of GAC model are the worst among all the test images. That is because the contour is initialized in the image center in this experiment, which cannot contain the objects of interest completely. That is to say, GAC model can detect the real object boundary only when the contour is initialized in specific locations. CV model and SLGS model can automatically detect all the boundaries in the image, regardless of the location of initial contour, so their F values are higher than GAC model. However, when the IR images contain severely local intensity inhomogeneity, global gray average information cannot drive the evolving curve stopping correctly. Under the circumstances, over segmentation and boundary leakage occur and F values decrease remarkably. As for LPF model and LoG & RSF model, they only consider local intensity information and ignore the global information. Therefore, they are prone to local minimum, leading to the result that the object is divided into many blocks mistakenly and the average F values of these two models are mediocre in our comparative experiments. LIC model which combines ACMs with bias field estimation is another strategy to cope with intensity inhomogeneity, but it is only suitable for images with simple background. So, the F values of LIC model are sometimes unsatisfactory because the curves are unable to converge in complex background regions. Compared to the above-mentioned ACMs, our method gets the largest average F values and its F value of each test images is acceptable, i.e., the proposed method can detect real object boundary more accurately and deal with inhomogeneous region more effectively.

Table.2 Statistical results of F values.

| IR image | GAC | CV | SLGS | LPF | LoG & RSF | LIC | Ours |
|---|---|---|---|---|---|---|---|
| 1 | 0.2031 | **0.9599** | **0.9013** | 0.7690 | 0.7044 | 0.5395 | **0.9535** |
| 2 | 0.4729 | **0.9540** | **0.9682** | 0.7137 | 0.6903 | 0.8824 | **0.9793** |
| 3 | 0.2058 | **0.9175** | 0.9172 | 0.5164 | 0.6318 | **0.9632** | **0.9897** |
| 4 | 0.2976 | **0.9646** | **0.9614** | 0.8139 | 0.7807 | 0.9240 | **0.9669** |
| 5 | 0.1003 | 0.9443 | **0.9450** | 0.7312 | 0.7653 | **0.9682** | **0.9868** |
| 6 | 0 | **0.9807** | 0.9656 | 0.6171 | 0.6816 | **0.9771** | **0.9792** |
| 7 | 0.0641 | **0.9673** | **0.9711** | 0.0777 | 0.0924 | 0.9391 | **0.9732** |
| 8 | 0.0536 | **0.9355** | **0.9658** | 0.0670 | 0.0879 | 0.8938 | **0.9698** |
| 9 | 0.5889 | **0.9183** | **0.9254** | 0.8869 | 0.7736 | 0.8751 | **0.9277** |
| 10 | 0.1003 | **0.9538** | 0.9496 | 0.6147 | 0.6618 | **0.9682** | **0.9843** |
| 11 | 0.2695 | **0.9668** | **0.9722** | 0.6862 | 0.6819 | 0.9475 | **0.9836** |
| 12 | 0.2104 | **0.9536** | 0.9512 | 0.6895 | 0.7381 | **0.9706** | **0.9834** |
| 13 | 0.0356 | **0.9267** | **0.9395** | 0.6632 | 0.6817 | 0.7557 | **0.9509** |
| 14 | 0 | **0.9676** | **0.9630** | 0.7505 | 0.7218 | 0.6587 | **0.9792** |
| 15 | 0.2975 | 0.3414 | **0.9198** | **0.8771** | 0.8430 | 0.3236 | **0.9347** |
| 16 | 0.3111 | **0.8342** | **0.9149** | 0.6185 | 0.5549 | 0.5827 | **0.9450** |
| Ave. | 0.2007 | **0.9054** | **0.9457** | 0.6308 | 0.6307 | 0.8231 | **0.9680** |

**4.2.3 Comparison of running time**

In addition to F value, running time is also employed as a metric to evaluate our algorithm. In the same experimental platform, the running time of different models is reported in Table.3 and the three top smallest ACMs have been marked in red, green and blue, respectively.

As shown in Table.3, we can find that GAC model and CV model are time-consuming due to their property of not easy to convergence. The average running time of GAC model and CV model is approximately 6 times and 2.5 times longer than that of our algorithm. LPF model is a kind of fast image segmentation model because it computes the pre-fitting functions before the evolution of curve and the pre-fitting functions do not need to update in each iteration. However, LPF model needs to minimize the local energy function which needs plenty of time. Although the LoG & RSF model imposes optimized LoG operator to detect the edge more effectively, but it still costs a lot of time to converge as the LPF model. As a result, the execution efficiencies of LPF and LoG & RSF model are moderate in the whole comparison. LIC model achieves satisfactory running time in some test images with simple background but performs poorly in others. Therefore, its calculation time is directly decided by the characteristics of image. SLGS model gets the fastest execution speed because it only needs to calculate the global intensity information to guide curve evolution, so its running efficiency is the highest. As for our method, the average running time is only more than SLGS model (approximately 3 seconds longer), because we introduce the local feature to reform the SPF function. We plan to transplant our algorithm to higher-speed computing platform and achieve real-time running in the future work. Considering the accuracy of segmentation results mentioned above, we can draw a conclusion that our method can obtain desirable results and acceptable execution time in IR image segmentation.

Table.3 Comparison of running time (in seconds)

| IR image | GAC | CV | SLGS | LPF | LoG & RSF | LIC | Ours |
|---|---|---|---|---|---|---|---|
| 1 | 18.1139 | **13.5297** | **3.3378** | 20.7844 | 20.4282 | 14.9276 | **6.0533** |
| 2 | 27.3372 | 12.3852 | **2.9496** | **7.3727** | 12.1505 | 9.1231 | **3.6548** |
| 3 | 27.1972 | **9.1124** | **2.9798** | 19.5277 | 24.8085 | 14.2686 | **3.0182** |
| 4 | 70.0298 | **10.1436** | 20.5524 | 21.2821 | 50.8191 | **13.5070** | **12.6598** |
| 5 | 88.6933 | **12.1157** | **9.8170** | 13.8495 | 22.7730 | 44.8938 | **13.8233** |
| 6 | 40.8549 | 33.9340 | **4.1179** | 34.8879 | 23.7596 | **12.2659** | **5.8662** |
| 7 | 87.3778 | 75.4667 | **13.6767** | 37.7720 | 33.9261 | **12.9943** | **18.2724** |
| 8 | 87.2557 | 105.3378 | **14.6998** | 75.2106 | 33.8854 | **17.1493** | **19.0991** |
| 9 | 28.7005 | **4.9718** | **2.6225** | **5.6645** | 12.0644 | 16.4423 | 8.8490 |
| 10 | 99.0442 | 28.0527 | **11.5100** | 44.3073 | 40.4406 | **27.6962** | **22.1138** |
| 11 | 38.8612 | 15.9982 | **4.2488** | 12.3188 | 30.9024 | **7.5761** | **8.6286** |
| 12 | 99.5036 | 18.9900 | **5.5146** | 15.7955 | 25.1829 | **9.6633** | **7.2771** |
| 13 | 174.6654 | 29.7931 | **7.3969** | 35.9777 | 51.1787 | **29.0758** | **9.7427** |
| 14 | 105.9294 | **19.8835** | **6.9861** | 59.6416 | 61.7432 | 25.5123 | **15.4654** |
| 15 | 47.5315 | 18.7070 | **5.1385** | **14.7697** | 18.8393 | 43.9738 | **6.8167** |
| 16 | 27.2223 | **5.5518** | **2.8613** | 6.8161 | 11.1171 | **3.3074** | 6.6630 |
| Ave. | 66.7699 | 25.8733 | **7.4006** | 26.6236 | 29.6262 | **18.8985** | **10.5002** |

## 5. Conclusion

In this paper, to segment IR images with intensity inhomogeneity effectively, we design a special SPF function which contains not only global information but also local feature information. Firstly, the global term computed by global gray average information guides the evolving curve to pass through the background region quickly and prevents the evolving curve falling into local minimum. The local term calculated by local entropy, local standard deviation and gradient information provides more local properties to determine the evolution direction more precisely in real boundary region. Then, the global term and local term are combined by an adaptive weight coefficient based on local range information to construct the complete SPF function. Next, LSF is rewritten by the new SPF function and meanwhile the Gaussian filter is utilized to re-initialize LSF. By finite iterations, we finally obtain the contour of object correspond to zero LSF. Both qualitative and quantitative experiments are implemented to compare our method with other ACMs, proving that our method outperforms the comparing ones in terms of precision rate and overlapping rate. To further decrease the running time, we plan to execute our proposed algorithm in higher-speed computing platform in the future.

**Acknowledgement**
This research was supported by National Natural Science Foundation of China (No. 62001234), Natural Science Foundation of Jiangsu Province (No. BK20200487), China Postdoctoral Science Foundation (No. 2020M681597) and Postdoctoral Science Foundation of Jiangsu Province (No. 2020Z051).